\documentclass{article} 
\usepackage{iclr2026_conference,times}
\usepackage{graphicx}
\usepackage{subcaption}
\usepackage{booktabs} 
\usepackage{amsmath}
\usepackage{wrapfig}
\usepackage[linesnumbered,ruled,vlined]{algorithm2e}
\usepackage{tikz}
\usepackage{pgfplots}
\pgfplotsset{compat=1.18}
\usepackage{url}
\usepackage{listings}
\usepackage{multirow}
\usepackage{hyperref}

\title{Vision-Guided Iterative Refinement for\\ Frontend Code Generation}

\author{Hannah Sansford\thanks{Work done while at Amazon.} \\
School of Mathematics, University of Bristol, UK \\
\texttt{hannah.sansford@bristol.ac.uk} \\
\AND
Derek H.~C. Law, Wei Liu, Abhishek Tripathi, Niresh Agarwal, Gerrit J.~J. van den Burg \\
Amazon AGI \\
\texttt{\{lawdere,weliuz,dtriabhi,nirea,gvdburg\}@amazon.com}
}

\iclrfinalcopy
\begin{document}

\maketitle

\begin{abstract}
Code generation with large language models often relies on multi-stage human-in-the-loop refinement, which is effective but very costly — particularly in domains such as frontend web development where the solution quality depends on rendered visual output. We present a fully automated critic-in-the-loop framework in which a vision-language model serves as a visual critic that provides structured feedback on rendered webpages to guide iterative refinement of generated code. Across real-world user requests from the WebDev Arena dataset, this approach yields consistent improvements in solution quality, achieving up to 17.8\% increase in performance over three refinement cycles. Next, we investigate parameter-efficient fine-tuning using LoRA to understand whether the improvements provided by the critic can be internalized by the code-generating LLM. Fine-tuning achieves 25\% of the gains from the best critic-in-the-loop solution without a significant increase in token counts. Our findings indicate that automated, VLM-based critique of frontend code generation leads to significantly higher quality solutions than can be achieved through a single LLM inference pass, and highlight the importance of iterative refinement for the complex visual outputs associated with web development.
\end{abstract}

\section{Introduction}

Modern code-generation systems using large language models (LLMs) frequently rely on multi-stage human-in-the-loop workflows: given a user query a model produces an initial solution, a human analyzes and critiques the code and its execution output, and the model is prompted to revise its output accordingly. While effective, such critic-guided refinement pipelines are costly and difficult to scale due to their reliance on repeated human inspection. This reliance on feedback becomes more important when code produces rich visual outputs, which cannot be reliably assessed until after execution and rendering. Frontend web development exemplifies this challenge, as correctness depends not only on functional behavior but also on visual layout, styling, and user interface fidelity.

We present a critic-in-the-loop (CITL) system for natural language-prompted frontend code generation tasks by introducing a vision-language model (VLM) as a visual critic. Our approach contrasts with previous work such as \citet{gou2024critic}, in which an LLM uses external tools such as search engines to inform a critique, since the visual critic only observes the rendering of the generated frontend code. Based on this rendering the VLM provides structured visual feedback that is passed to a secondary critic along with the generated code, in order to produce a consolidated critique that drives iterative self-refinement. We quantify the resulting improvements using a VLM-as-a-Judge evaluation framework \citep{chen2024mllm} -- which is validated against human-annotated preference data -- and analyze the trade-off between quality and token usage. 

Prior work has explored vision-grounded frontend code refinement on image-to-code settings, including chart replication \citep{li2025metal}, webpage replication \citep{si2025design2code} and sketch-to-webpage generation \citep{li2025sketch2code}. In contrast, we focus on the fundamental task of developing a web UI based on a user query, which remains underexplored in the literature. Moreover, we highlight a key limitation of such pipelines: while iterative visual critiques consistently improve solution quality, they incur substantial computational overhead due to repeated rounds of code generation, rendering, and critique.

Motivated by this trade-off we investigate whether a code-generating LLM can \textit{internalize} the improvements achieved through critic-in-the-loop refinement and produce higher-quality code in a single pass. More broadly, this draws on the idea of knowledge distillation, which transfers the behavior of a more expensive teacher into a cheaper student \citep{hinton2015distilling,kim2016sequence}. In our setting, the teacher is an iterative, multimodal critic-in-the-loop pipeline, and we employ parameter-efficient LoRA fine-tuning \citep{hu2022lora} to distill the benefits of VLM-provided visual feedback into the weights of the code-generating LLM. This avoids the high token usage and latency of iterative refinement while retaining performance gains.

We summarize our contributions as follows:
\begin{enumerate}
    \item We present a VLM-based critic in the loop pipeline for Web UI code generation and demonstrate its performance in the self-refinement and student-teacher settings on a dataset of real-world user queries from WebDev Arena \citep{vichare2025webdev}.
    \item We introduce a novel VLM-as-a-Judge approach for evaluating LLM-generated Web UIs that considers task accomplishment, aesthetics, and code quality, and validate that it aligns with human preference votes.
    \item We fine-tune the code generator component of our CITL system using LoRA on an open-weight model to quantify the extent to which critic feedback can be internalized by the model.  
\end{enumerate}
In Section~\ref{sec:related_work} we consider three categories of previous work related to ours. Section~\ref{sec:methodology} introduces our methodology, including our critic-in-the-loop system design and our approach to evaluation and LoRA fine-tuning. Experiments to validate our VLM-as-a-Judge are presented in Section~\ref{sec:experiments}, along with experiments with our CITL pipeline on three code-generating LLMs and the LoRA fine-tuning approach. We conclude the paper and discuss limitations and future work in Section~\ref{sec:discussion}.

\begin{figure}[t]
\centering

\begin{minipage}[c]{0.58\linewidth}
    \centering
    \includegraphics[width=.9\linewidth]{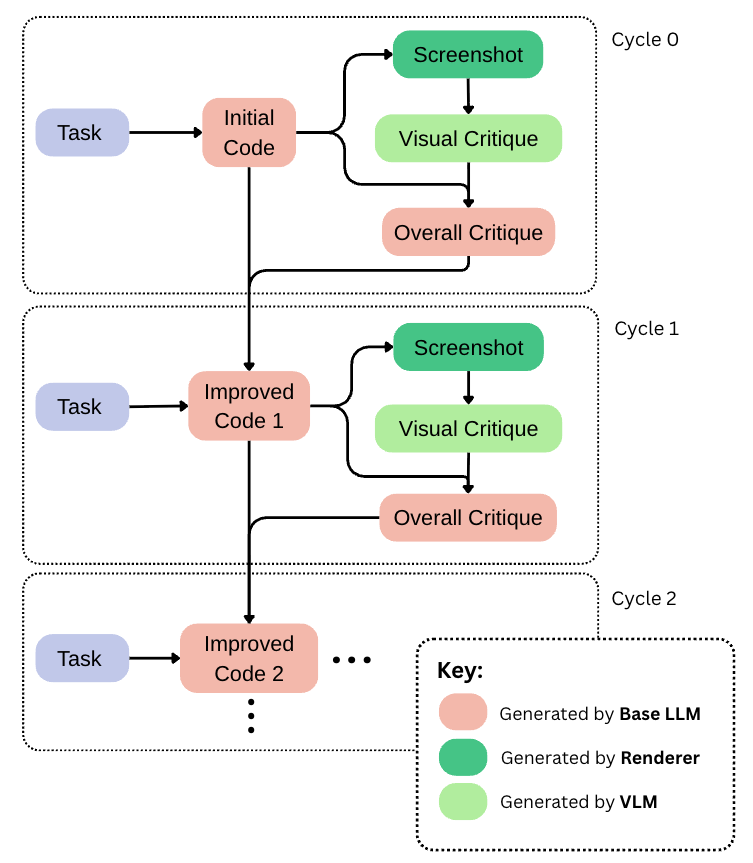}
\end{minipage}
\hfill
\begin{minipage}[c]{0.41\linewidth}
    \centering
    \includegraphics[width=.9\linewidth]{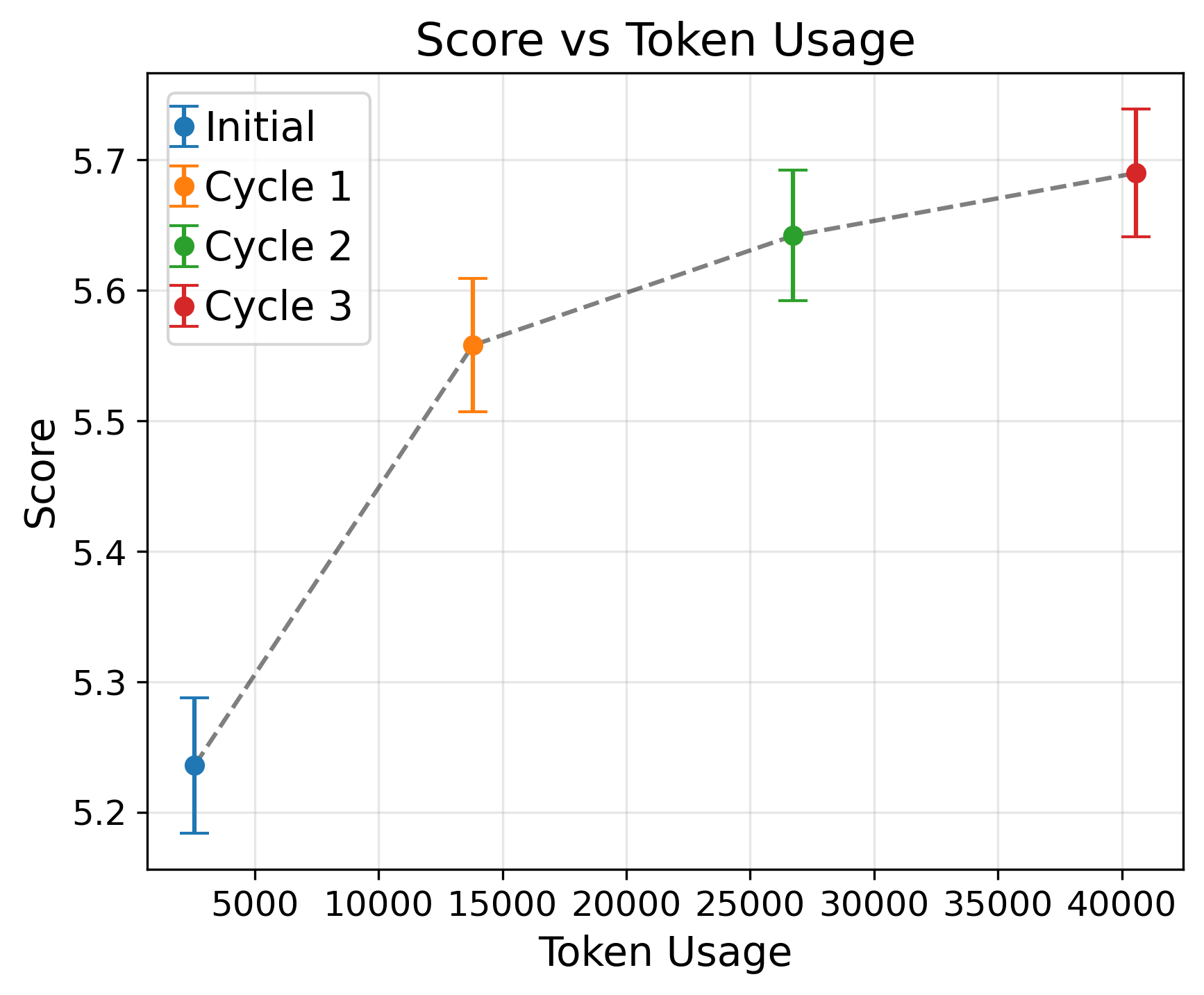}

    \vspace{0.6em}

    \includegraphics[width=.9\linewidth]{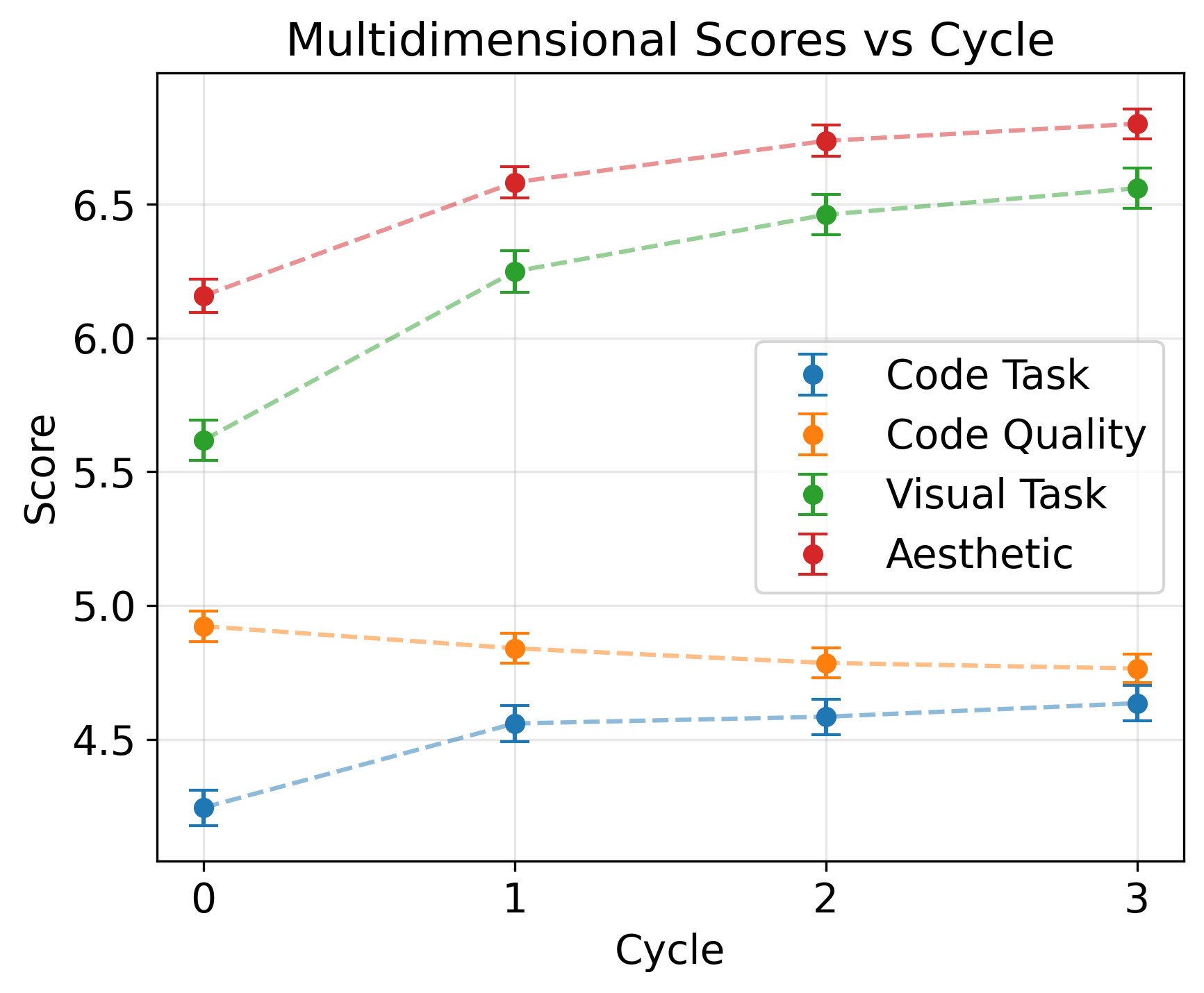}
\end{minipage}

\caption{%
\textbf{Left:} Overview of the critic-in-the-loop (CITL) pipeline for web code generation. 
\textbf{Right (top):} Overall score versus token usage, illustrating the quality–efficiency trade-off across refinement cycles. 
\textbf{Right (bottom):} Breakdown of VLM-judged performance across four evaluation dimensions.
Performance figures are based on the Distill-Qwen-14B generator model.
}
\label{fig:citl_and_results}
\end{figure}

\section{Related Work}
\label{sec:related_work}
In this section we discuss related work on vision-grounded code improvement, the use of a VLM-as-a-Judge, and knowledge distillation.

\paragraph{Vision-grounded code refinement.}
Building on early research showing that LLMs can produce useful feedback and iteratively improve outputs \citep{madaan2023self}, recent work has explored using multimodal models as visual critics to ground code generation and refinement. ReLook \citep{li2025relook} introduces an agentic, vision-grounded reinforcement learning framework in which a multimodal LLM serves as a critic, providing visual feedback from rendered webpages to guide policy learning for interactive web-coding agents. For chart replication, METAL \citep{li2025metal} employs collaborating VLM-powered agents and shows that separating modalities during critique leads to strong self-correction and test-time scaling benefits. 
For frontend code generation based on a visual design,
Design2Code \citep{si2025design2code} explores visual self-revision prompting by conditioning the model on screenshots of both a reference webpage and its own rendered output, and prompting it to revise the initial implementation to better match the target design.

\paragraph{VLM-as-a-Judge.}
Recent work has shown that LLMs can serve as effective automated judges for evaluating generative model outputs, providing scalable alternatives to human evaluation when carefully validated. Many prior studies have explored this paradigm in text-only settings, analyzing agreement with human preferences and identifying sources of bias in LLM-based evaluation \citep{zheng2023judging, chiang2024chatbot}. More recently, MLLM-as-a-Judge \citep{chen2024mllm} extends this framework to multimodal settings, systematically assessing VLMs as judges across visual-language benchmarks and demonstrating strong correlation with human judgments under appropriate prompting and calibration. In this work, we adopt a VLM-as-a-Judge approach to evaluate the visual and code quality of generated webpages, and validate its reliability through comparison with human-annotated preference data.

\paragraph{Knowledge distillation.} Knowledge distillation was introduced as a student-teacher paradigm where a compact model is trained to match the softened output distributions of a more expressive teacher \citep{hinton2015distilling}. Subsequent work showed that distillation need not be limited to logits, and can instead operate at the level of entire generated sequences, enabling students to imitate the behavioral outputs of a stronger model in autoregressive settings \citep{kim2016sequence}. This idea has been successfully applied to LLMs, where distilled students approximate the generation patterns of larger teachers while retaining much of their performance \citep{sanh2019distilbert}. In parallel, parameter-efficient adaptation methods such as LoRA \citep{hu2022lora} provide a practical mechanism for internalizing such behaviors in large pretrained models without full fine-tuning. In addition to our CITL pipeline we consider distilling the \textit{behavior induced by critic-guided refinement}. Specifically, we explore LoRA fine-tuning on the best code solutions produced by an iterative, multimodal critic-in-the-loop process as a form of knowledge distillation.

\section{Methodology}
\label{sec:methodology}
We first introduce our vision-guided critic-in-the-loop pipeline for frontend code generation, followed by our approach to evaluation and LoRA fine-tuning.

\begin{figure}[h!]
\centering
\begin{minipage}[c]{0.48\linewidth}
\centering
\renewcommand{\arraystretch}{1.25}
\begin{tabular}{ll}
\hline
\textbf{Symbol} & \textbf{Meaning} \\
\hline
$G$ & Generator \\
$R$ & Renderer \\
$V$ & Visual Critic \\
$C$ & Code Critic \\
$I$ & Improver \\
$E$ & Evaluator \\
$x$ & Task \\
$t$ & Cycle / iteration index \\
$y_t$ & Generated code at cycle $t$ \\
$s_t$ & Evaluated score at cycle $t$ \\
$v_t$ & Visual critique at cycle $t$ \\
$c_t$ & Consolidated critique at cycle $t$ \\
\hline
\end{tabular}
\end{minipage}
\hfill
\begin{minipage}[c]{0.48\linewidth}
\centering
\begin{algorithm}[H]
\caption{Critic-in-the-Loop (CITL)}
\label{alg:critic-loop}
\KwIn{Task $x$, iteration budget $T$}
\KwOut{Best generated code $y_{\text{best}}$}

$y_0 \gets G(x)$\; 
$s_0 \gets E(x, y_0, R(y_0))$\; 
$y_{\text{best}} \gets y_0$\;
$s_{\text{best}} \gets s_0$\;

\For{$t = 0$ \KwTo $T-1$}{
    $v_t \gets V(x, R(y_t))$\;
    $c_t \gets C(x, y_t, v_t)$\;
    $y_{t+1} \gets I(x, y_t, c_t)$\;
    $s_{t+1} \gets E(x, y_{t+1}, R(y_{t+1}))$\;
    \If{$s_{t+1} \ge s_{\text{best}}$}{
        $y_{\text{best}} \gets y_{t+1}$, \quad $s_{\text{best}} \gets s_{t+1}$\;
    }
}%
\Return $y_{\text{best}}$\;
\end{algorithm}
\end{minipage}
\caption{Notation used in the paper (left) alongside the Critic-in-the-Loop (CITL) algorithm (right).}
\label{fig:notation_alg}
\end{figure}

\subsection{Critic-in-the-loop}

We formalize our critic-in-the-loop (CITL) pipeline for frontend code generation in Algorithm~\ref{alg:critic-loop} and illustrate the overall data flow in Figure~\ref{fig:citl_and_results}. A summary of the notation can be found in Figure~\ref{fig:notation_alg}. Prompt templates for each component are provided in Appendix~\ref{app:prompts}. We focus here on generating vanilla HTML code, but our system is equally applicable to frontend code that uses frameworks (such as React, Vue, etc). The generator $G$, code critic $C$, and improver $I$ are implemented using an LLM, while the visual critic $V$ and evaluator $E$ are implemented using a VLM. Below, we describe each stage of the pipeline.

\paragraph{Initial generation.}
Given a natural language task description $x$, the generator $G$ produces an initial frontend code implementation $y_0$. This implementation is rendered by $R$ to produce a visual representation of the webpage, which serves as the basis for all downstream evaluation and critique. We implement the renderer using Selenium\footnote{\url{https://www.selenium.dev/}}, but alternative browser automation tools are possible.

\paragraph{Visual critique.}
At iteration $t$, the visual critic $V$ inspects the rendered output $R(y_t)$ and analyzes how well it satisfies the task requirements from an end-user perspective. The resulting visual critique, $v_t$, captures issues related to layout, styling, visual hierarchy, spacing, and overall aesthetic coherence, while abstracting away from code-level details.

\paragraph{Code-level critique.} The code critic $C$ combines the task $x$, generated code $y_t$, and visual critique $v_t$ to produce an integrated critique, $c_t$, prioritizing issues identified by $V$ and translating visual deficiencies into actionable code-level recommendations. This two-stage design -- first assessing the rendered output, then reasoning about code -- was demonstrated to be more effective than a single unified critic by \citet{li2025metal}. By separating these responsibilities, refinements target user-visible failures rather than code issues.

\paragraph{Improvement and selection.}
The improver $I$ applies the critique $c_t$ to produce a revised implementation $y_{t+1}$. Each candidate solution is evaluated by $E$ using both the code and its rendered output, yielding a score $s_t$. Across iterations, the system retains the highest-scoring solution $y_{\text{best}}$.

\paragraph{Iterative refinement.}
This process is repeated for a fixed iteration budget $T$. While successive iterations typically improve visual fidelity and task alignment, the pipeline incurs increasing computational cost due to repeated generation, rendering, and critique. This trade-off motivates our distillation-based approach, which aims to internalize the benefits of critic-guided refinement into a single-pass generator.

We implement our CITL pipeline using Ray \citep{moritz2018ray} for efficient and asynchronous execution of the components.

\subsection{Evaluation}

Evaluating LLM-generated frontend code is an active area of research \citep{xiao2025code,li2025webdevjudge} and is complicated by the fact that evaluating frontend code involves analyzing both code level details and visual aesthetics. In line with our CITL pipeline design where visual and code critiques are separated, we implement a VLM-as-a-Judge (VLMaaJ) that scores both modalities individually. In Section~\ref{sec:experiments} we compare this approach against a VLMaaJ that considers both modalities simultaneously and that produces a single evaluation score (for prompts see Appendix~\ref{app:evaluator}).

Our multi-dimensional VLMaaJ assesses each generated solution $y_t$ along four complementary dimensions, scored on a 1-10 scale:
\begin{enumerate}
    \item Visual Task Accomplishment: how well the rendered output meets the task requirements.
    \item  Aesthetic Quality: visual appeal, layout, typography, color usage and modern design.
    \item Code Task Accomplishment: how well code implementation addresses task requirements.
    \item Code Quality: code structure, clarity, functionality and best practices.
\end{enumerate}
The first two dimensions are evaluated based solely on the user query and the rendered webpage, while the latter two are evaluated using only the user query and the generated frontend code. Evaluating task accomplishment in both the visual and the code domains ensures that functionality that is not visually apparent (such as pop ups or certain forms of interactivity) can still be considered by the judge. We compute an overall score by averaging across all four dimensions.

\subsection{LoRA Distillation}
\label{sec:distillation}

While critic-in-the-loop refinement can improve solution quality, it incurs substantial computational overhead due to repeated rounds of generation, rendering, and critique. To mitigate this cost, we investigate whether the improvements achieved through iterative visual critique can be internalized by the code-generating model and recovered in a single generation step. For each training task $x$, we run the full CITL pipeline and retain the highest-scoring solution $y_{\text{best}}$, producing a dataset of $(x, y_{\text{best}})$ pairs. We then fine-tune the base LLM with a parameter-efficient method (LoRA) to directly generate $y_{\text{best}}$ from $x$, without access to intermediate critiques or rendered outputs. Through this process, the model internalizes patterns and corrections that would otherwise require multiple rounds of visual feedback and regeneration.

\section{Experiments \& Results}
\label{sec:experiments}

We first discuss our experimental setup, including the dataset and models used. Next we describe how we validated the VLM-as-a-Judge for frontend code evaluation. Finally, we present our experiments on the critic-in-the-loop system using several code generating models, including the self-improvement setting. We analyze how the visual aesthetics and code quality improves over cycles, and provide a deep-dive into the content of the critiques.

\subsection{Experimental Setup}

\paragraph{Data.} We leverage the WebDev Arena dataset \citep{vichare2025webdev} as a source of real-world user queries for frontend code and human preference votes.\footnote{\url{https://huggingface.co/datasets/lmarena-ai/webdev-arena-preference-10k}} We filter the raw dataset to remove queries that do not request a website or web application, and filter out non-English requests for ease of use. From the remaining 4.4k queries we uniformly sample 2000 that we subsequently split into 1200 training tasks (for LoRA fine-tuning), 200 validation tasks, and 600 test tasks. 

\paragraph{Models.} Our system is flexible in the choice of model for the code generator, visual critic, and code critic. For our experiments we fix the visual critic to be Claude 4.5 Sonnet \citep{anthropic2025claude}. Furthermore, we ensure that the code generator and code critic are the same model, and for these models we evaluate DeepSeek-R1-Distill-Qwen-14B \citep[``Distill-Qwen-14B'';][]{guo2025deepseekr1}, Claude 4.5 Haiku, and Claude 4.5 Sonnet. The latter allows us to measure the self-improvement capability of  our system, as in that setting Claude 4.5 Sonnet is the code generator, code critic, and visual critic.

\paragraph{LoRA Fine-tuning.} 
We fine-tune the Distill-Qwen-14B model using LoRA to internalize the improvements achieved by CITL. Training uses the highest-scoring CITL outputs on the training dataset as targets, and leverages early stopping on the validation loss. LoRA is applied to the key linear projection matrices in the model's self-attention and feedforward layers, including the query, key, value, output, gating, and feedforward projections, with rank $r=16$, scaling factor $\alpha=32$, and dropout $0.1$. Fine-tuning minimizes the standard autoregressive language modeling loss over the refined code sequences, conditioning only on the original task description.

\subsection{Evaluation}

To assess the benefits of this multi-dimensional evaluation scheme, we compare it against a simpler single-dimensional alternative (see Appendix~\ref{subsub:single_dim}). In the single-dimensional setting, the VLM is prompted to return a single overall score between 1 and 10 for each solution, based jointly on the rendered output and the corresponding code. The prompt includes a descriptive rubric specifying the qualitative criteria associated with each discrete score.

\begin{table}[t]
\centering
\caption{Evaluation of agreement of a single-dimensional and a multi-dimensional VLM-as-a-Judge with human preferences, demonstrating stronger agreement for the multi-dimensional approach. \label{tab:judge_validation}}
\begin{tabular}{lrr}
\toprule
\textbf{Outcome} &
\textbf{Single-dimensional Judge} &
\textbf{Multi-dimensional Judge} \\
\midrule
Agreement & 48.5\% & 69.5\% \\
Draw & 28.5\% & 8.5\% \\
Disagreement & 23.0\% & 22.0\% \\
\bottomrule
\end{tabular}
\end{table}

We evaluate both single-dimensional and multi-dimensional judge implementations by measuring their agreement with human judgments from the WebDev Arena preference dataset, which contains pairwise comparisons annotated with a human preference vote. We sample 200 tasks with a clear human preference, excluding pairs labeled as ties. If the judge assigns a higher score to the human-preferred solution, we count this as \emph{agreement}, and if the preferred solution receives a lower score it is counted as \emph{disagreement}. Equal scores are considered a \emph{draw}.

Table~\ref{tab:judge_validation} reports the resulting agreement statistics, which demonstrate that our multi-dimensional judge exhibits 43.3\% higher agreement with human preferences than the single-dimensional alternative. Moreover, it is significantly more discriminative, resulting in draws for only 8.5\% of tasks compared to 28.5\% under single-dimensional scoring. While coarser scoring schemes naturally induce more ties, recent work has shown that increasing the granularity of single-score judgments can introduce additional noise and reduce alignment with human preference \citep{li2026grading}. Our results align with recent findings in \citet{li2025webdevjudge}, where agreement of 69.77\% is reported for a different judge implementation and a different subset of the WebDev Arena dataset. Based on these results we adopt the multi-dimensional evaluation scheme for all experiments.

\subsection{Main Results}

\begin{table}[tb]
\centering
\caption{Performance comparison across different generator models and critic-in-the-loop cycles, as measured by the multidimensional VLM judge. Also shown is the performance of refinement without a critic and the performance of the Distill-Qwen-14B model after LoRA fine-tuning. The best solution obtained during critic in the loop is marked by ``CITL: Best''.}
\label{tab:main_results}
\small
\begin{tabular}{llcrr}
\toprule
\textbf{Generator} & \textbf{Method} & \textbf{Score} & \textbf{Gain (\%)} & \textbf{Tokens} \\
\midrule
\multirow{7}{*}{Distill-Qwen-14B} 
& Initial code & $5.236$ ${\scriptstyle \pm 0.052}$ & -- & 2,571 \\
& CITL: Cycle 1 & $5.558$ ${\scriptstyle \pm 0.051}$ & \textcolor{green!50!black}{+6.2} & 15,067 \\
& CITL: Cycle 2 & $5.642$ ${\scriptstyle \pm 0.050}$ & \textcolor{green!50!black}{+7.8} & 29,853 \\
& CITL: Cycle 3 & $5.690$ ${\scriptstyle \pm 0.049}$ & \textcolor{green!50!black}{+8.7} & 45,808 \\
& CITL: Best & \underline{$6.165$} ${\scriptstyle \pm 0.043}$ & \textcolor{green!50!black}{+17.8} & 23,662 \\
& Refine w/o Critic & $5.317$ ${\scriptstyle \pm 0.053}$ & \textcolor{green!50!black}{+1.5} & 7{,}209 \\
& LoRA fine-tuning & $5.467$ ${\scriptstyle \pm 0.055}$ & \textcolor{green!50!black}{+4.4} & 2{,}583 \\
\midrule
\multirow{5}{*}{Haiku 4.5} 
& Initial code & $7.436$ ${\scriptstyle \pm 0.042}$ & -- & 6,753 \\
& CITL: Cycle 1 & $7.706$ ${\scriptstyle \pm 0.035}$ & \textcolor{green!50!black}{+3.6} & 37,604 \\
& CITL: Cycle 2 & $7.820$ ${\scriptstyle \pm 0.033}$ & \textcolor{green!50!black}{+5.2} & 77,787 \\
& CITL: Cycle 3 & $7.862$ ${\scriptstyle \pm 0.037}$ & \textcolor{green!50!black}{+5.7} & 124,599 \\
& CITL: Best & \underline{$8.160$} ${\scriptstyle \pm 0.030}$ & \textcolor{green!50!black}{+9.8} & 65,005 \\
\midrule
\multirow{5}{*}{Sonnet 4.5} 
& Initial code & $7.584$ ${\scriptstyle \pm 0.040}$ & -- & 6,446 \\
& CITL: Cycle 1 & $7.958$ ${\scriptstyle \pm 0.034}$ & \textcolor{green!50!black}{+4.9} & 37,642 \\
& CITL: Cycle 2 & $8.021$ ${\scriptstyle \pm 0.035}$ & \textcolor{green!50!black}{+5.8} & 80,472 \\
& CITL: Cycle 3 & $8.048$ ${\scriptstyle \pm 0.042}$ & \textcolor{green!50!black}{+6.1} & 131,550 \\
& CITL: Best & \underline{$8.402$} ${\scriptstyle \pm 0.025}$ & \textcolor{green!50!black}{+10.8} & 73,215 \\
\bottomrule
\end{tabular}
\end{table}

The results from our critic-in-the-loop experiments are reported in Table~\ref{tab:main_results} and shown visually in Figure~\ref{fig:citl_and_results} for the Distill-Qwen-14B model. Scores are reported as the mean over all examples in a given cycle. We observe consistent improvement through the cycles for each code generator model, demonstrating the benefit of our approach. Looking at the best solution obtained through the cycles, we find a performance improvement of 17.8\% with the Distill-Qwen-14B model as code generator, and 10.8\% in the self-improvement setting with Claude 4.5 Sonnet. The self-improvement setting in particular benefits from the critic-in-the-loop approach, as the best solution is obtained after one or more cycles for 86\% of tasks.

To isolate the effect of the critic-in-the-loop approach, we also consider a setting in which the code generator LLM is prompted to refine the initial code without any visual or code feedback (``Refine w/o Critic''; prompt provided in Appendix~\ref{app:refine_wo_crtic}). On the Distill-Qwen-14B model this approach produces only a 1.5\% improvement, confirming that critic-guided refinement is the primary driver of performance gains we observe.

Applying our CITL system in the knowledge distillation setting on the Distill-Qwen-14B model, we observe that LoRA fine-tuning captures 24.7\% of the improvements observed in the best CITL solution and 50.6\% of the third cycle, with comparative token usage to the initial code generation. This demonstrates that fine-tuning the model can internalize some of the iterative critic guidance, achieving substantial quality gains in a single-pass generation. However, it also shows lasting benefits of the critic in the loop approach.

\subsection{Multi-dimensional Breakdown}

\begin{table}[t]
\centering
\caption{Breakdown of performance across four evaluation dimensions for critic-in-the-loop refinement cycles and the two baselines, using the Distill-Qwen-14B model. Percentages indicate relative change compared to the initial code baseline.\label{tab:score_breakdown}}
\small
\begin{tabular}{lcccc}
\toprule
\textbf{Method} &
\begin{tabular}[c]{@{}c@{}}\textbf{Code Task}\\\textbf{Accomplishment}\end{tabular} &
\begin{tabular}[c]{@{}c@{}}\textbf{Code}\\\textbf{Quality}\end{tabular} &
\begin{tabular}[c]{@{}c@{}}\textbf{Visual Task}\\\textbf{ Accomplishment}\end{tabular} &
\begin{tabular}[c]{@{}c@{}}\textbf{Aesthetic}\\\textbf{Quality}\end{tabular} \\
\midrule
Initial code
& 4.244 \phantom{\textcolor{green!50!black}{(+0.0\%)}}
& 4.923 \phantom{\textcolor{green!50!black}{(+0.0\%)}}
& 5.618 \phantom{\textcolor{green!50!black}{(+00.0\%)}}
& 6.158 \phantom{\textcolor{green!50!black}{(+0.0\%)}} \\

CITL: Cycle 1
& 4.559 \textcolor{green!50!black}{(+7.4\%)}
& 4.840 \textcolor{red!70!black}{(-1.7\%)}
& 6.249 \textcolor{green!50!black}{(+11.2\%)}
& 6.582 \textcolor{green!50!black}{(+6.9\%)} \\

CITL: Cycle 2
& 4.584 \textcolor{green!50!black}{(+8.0\%)}
& 4.785 \textcolor{red!70!black}{(-2.8\%)}
& 6.461 \textcolor{green!50!black}{(+15.0\%)}
& 6.737 \textcolor{green!50!black}{(+9.4\%)} \\

CITL: Cycle 3
& 4.635 \textcolor{green!50!black}{(+9.2\%)}
& 4.764 \textcolor{red!70!black}{(-3.2\%)}
& 6.561 \textcolor{green!50!black}{(+16.8\%)}
& 6.801 \textcolor{green!50!black}{(+10.4\%)} \\

CITL: Best
& 4.998 \textcolor{green!50!black}{(+17.8\%)}
& 5.300 \textcolor{green!50!black}{(+7.6\%)}
& 7.094 \textcolor{green!50!black}{(+26.3\%)}
& 7.269 \textcolor{green!50!black}{(+18.0\%)} \\
\midrule
Refine w/o Critic
& 4.296 \textcolor{green!50!black}{(+1.2\%)}
& 4.942 \textcolor{green!50!black}{(+0.4\%)}
& 5.707 \textcolor{green!50!black}{(+1.6\%)}
& 6.321 \textcolor{green!50!black}{(+2.6\%)} \\
LoRA Fine-tuning
& 4.481 \textcolor{green!50!black}{(+5.6\%)}
& 5.099 \textcolor{green!50!black}{(+3.6\%)}
& 5.782 \textcolor{green!50!black}{(+2.9\%)}
& 6.506 \textcolor{green!50!black}{(+5.7\%)} \\
\bottomrule
\end{tabular}
\end{table}

Table~\ref{tab:score_breakdown} provides a multi-dimensional breakdown of performance across code and visual-level evaluation criteria for the Distill-Qwen-14B model (see Appendix~\ref{app:additional_results} for other models). Critic-in-the-loop refinement consistently leads to gains in visual task accomplishment and aesthetic quality, with improvements increasing monotonically across cycles. This reflects the benefit of a visual critic, which targets layout, visual hierarchy, and other user-visible deficiencies in the rendered output. Notably, the largest improvements are observed in visual task accomplishment, which is particularly relevant for Web UI generation. By contrast, code quality exhibits a slight decline across CITL cycles, suggesting that iterative refinement prioritizes perceptual and functional correctness over code-level cleanliness or style. This behavior is consistent with our system design: the code critic is prompted to reason about the implementation primarily through the lens of the visual feedback, rather than explicitly optimizing for code quality.

Refinement without critique produces only marginal improvements across all dimensions, indicating that unguided self-refinement is insufficient to substantially improve either visual or functional outcomes. The LoRA-distilled model, by contrast, recovers a substantial portion of the gains in aesthetic quality and code task accomplishment while also improving code quality relative to the base model. This suggests that distillation internalizes the outcomes of critic-guided refinement without inheriting its tendency to introduce incremental code-level complexity, enabling a more balanced trade-off between visual performance and code quality in a single-pass generation. The smaller gains in visual task accomplishment indicate that the distilled signal primarily captures general aesthetic and structural improvements rather than fine-grained task-specific visual requirements.

\subsection{LLM Critiques}
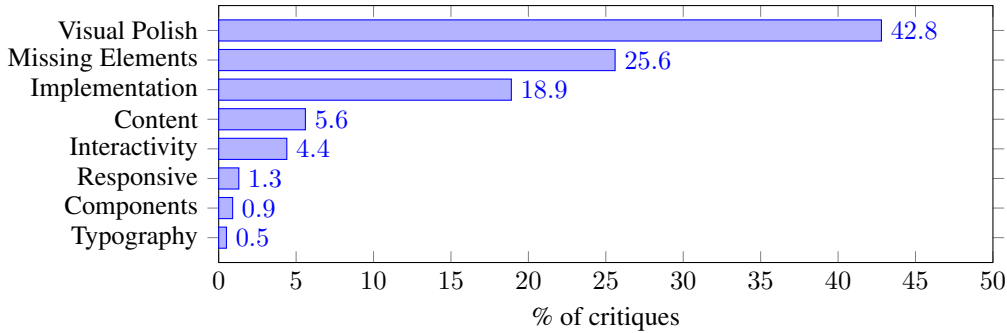
\begin{figure}[tb]
\centering
\begin{tikzpicture}
\begin{axis}[
    width=.85\textwidth,
    height=5cm,
    xbar,
    xmin=0,
    xmax=50,
    xlabel={\% of critiques},
    symbolic y coords={Typography, Components, Responsive, Interactivity, Content, Implementation, Missing Elements, Visual Polish},
    ytick=data,
    nodes near coords,
    nodes near coords align={horizontal},
    bar width=8pt,
    enlarge y limits=0.12,
]
\addplot coordinates {
    (0.5,Typography)
    (0.9,Components)
    (1.3,Responsive)
    (4.4,Interactivity)
    (5.6,Content)
    (18.9,Implementation)
    (25.6,Missing Elements)
    (42.8,Visual Polish)
};
\end{axis}
\end{tikzpicture}
\caption{Distribution of critique categories. \textit{Visual Polish}: spacing, alignment, refinement. \textit{Missing Elements}: incomplete sections or functionality. \textit{Implementation}: code quality, accessibility issues. \textit{Content}: hierarchy and organization. \textit{Interactivity}: hover states, feedback. \textit{Responsive}: mobile adaptation. \textit{Components}: UI element consistency. \textit{Typography}: fonts, sizing, readability. \label{fig:critique_dist}}
\end{figure}

We collected 7,300 critiques (based on visual and code critique) based on WebDev Arena tasks and classify the critiques into 8 high-level categories using Claude 4.5 Sonnet (see \ref{app:critique_classification} for the prompt used). Figure \ref{fig:critique_dist} shows that majority of the critiques are focused on visual aesthetics and functionality. These findings highlight that an effective CITL system for Web UIs must possess multimodal capabilities. Specifically, the model needs to analyze both the underlying code to detect structural completeness and semantic issues, and visual representations like screenshots to evaluate visual hierarchy, and overall polish. Therefore, a multimodal approach combining code understanding with visual perception is essential for accurately differentiating good webpage designs from poor ones.

\section{Discussion}
\label{sec:discussion}

We introduced a fully automated critic-in-the-loop pipeline that uses a VLM to provide visual feedback on rendered webpages, and showed that iterative critiques improve visual fidelity across three code generating models. We demonstrated that a significant part of this improvement can be internalized by the model through fine-tuning, but that benefits of the CITL approach remain. While our system focused on vision-grounded critiques for HTML frontend code, it can readily be applied to using web frameworks such as React and Vue, and can be extended to other code generation for visual outputs, such as SVG graphics, simply by modifying the system prompts.

Despite these promising results, some known limitations exist. First, our evaluation relies heavily on a VLM-as-a-Judge, which can exhibit biases or other deficiencies (e.g., over-/under-weighting certain aesthetics, missing accessibility issues). Although we validated VLM judgments against human preferences, broader human evaluation can be valuable to fully characterize alignment and failure modes. Second, we do not compare CITL (or the LoRA student) against a ceiling established by human critiques and corrections. Human experts may provide richer, contextual, and judgment-sensitive feedback that could enable higher-quality refinements than our automated VLM critic. 

There are several natural directions to extend this work. Distilling intermediate signals (visual critiques, structured edit suggestions, or saliency maps) in addition to final outputs may enable the model to better internalize the refinement \textit{process} rather than only its outcome. Similarly, understanding the cause of performance regressions and the diminishing returns of the critique can inspire novel approaches to critic-guided code generation. Regardless, our work convincingly demonstrates the lasting benefit of the CITL approach, both in the self-improvement and student-teacher settings.

\ificlrfinal
\section*{Acknowledgements}
The authors would like to thank Murat Sensoy, Iknoor Singh, and Zhunxuan Wang for helpful comments on the project.
\fi

\bibliography{bibliography}
\bibliographystyle{iclr2026_conference}

\appendix

\section{Additional Results}
\label{app:additional_results}

\begin{table}[h]
\centering
\caption{Component score by cycle for the Haiku 4.5 and Sonnet 4.5 generator models.}
\small
\begin{tabular}{llcccc}
\toprule
\textbf{Generator} &
\textbf{Method} &
\begin{tabular}[c]{@{}c@{}}\textbf{Code Task}\\\textbf{Accomplishment}\end{tabular} &
\begin{tabular}[c]{@{}c@{}}\textbf{Code}\\\textbf{Quality}\end{tabular} &
\begin{tabular}[c]{@{}c@{}}\textbf{Visual Task}\\\textbf{ Accomplishment}\end{tabular} &
\begin{tabular}[c]{@{}c@{}}\textbf{Aesthetic}\\\textbf{Quality}\end{tabular} \\
\midrule
\multirow{4}{*}{Haiku 4.5} 
& Initial code
& 7.110 \phantom{\textcolor{green!50!black}{+0.0\%}}
& 7.071 \phantom{\textcolor{green!50!black}{+0.0\%}}
& 7.804 \phantom{\textcolor{green!50!black}{+0.0\%}}
& 7.758 \phantom{\textcolor{green!50!black}{+0.0\%}} \\

& CITL: Cycle 1
& 7.571 \textcolor{green!50!black}{+6.5\%}
& 7.369 \textcolor{green!50!black}{+4.2\%}
& 8.002 \textcolor{green!50!black}{+2.5\%}
& 7.882 \textcolor{green!50!black}{+1.6\%} \\

& CITL: Cycle 2
& 7.707 \textcolor{green!50!black}{+8.4\%}
& 7.529 \textcolor{green!50!black}{+6.5\%}
& 8.082 \textcolor{green!50!black}{+3.6\%}
& 7.960 \textcolor{green!50!black}{+2.6\%} \\

& CITL: Cycle 3
& 7.782 \textcolor{green!50!black}{+9.5\%}
& 7.560 \textcolor{green!50!black}{+6.9\%}
& 8.135 \textcolor{green!50!black}{+4.2\%}
& 7.970 \textcolor{green!50!black}{+2.7\%} \\

& CITL: Best
& 7.995 \textcolor{green!50!black}{+12.4\%}
& 7.794 \textcolor{green!50!black}{+10.2\%}
& 8.522 \textcolor{green!50!black}{+9.2\%}
& 8.330 \textcolor{green!50!black}{+7.4\%} \\

\midrule
\multirow{4}{*}{Sonnet 4.5}
& Initial code
& 7.346 \phantom{\textcolor{green!50!black}{+0.0\%}}
& 7.192 \phantom{\textcolor{green!50!black}{+0.0\%}}
& 8.106 \phantom{\textcolor{green!50!black}{+0.0\%}}
& 7.690 \phantom{\textcolor{green!50!black}{+0.0\%}} \\

& CITL: Cycle 1
& 7.974 \textcolor{green!50!black}{+8.5\%}
& 7.660 \textcolor{green!50!black}{+6.5\%}
& 8.226 \textcolor{green!50!black}{+1.5\%}
& 7.974 \textcolor{green!50!black}{+3.7\%} \\

& CITL: Cycle 2
& 8.056 \textcolor{green!50!black}{+9.7\%}
& 7.683 \textcolor{green!50!black}{+6.8\%}
& 8.251 \textcolor{green!50!black}{+1.8\%}
& 8.093 \textcolor{green!50!black}{+5.2\%} \\

& CITL: Cycle 3
& 8.129 \textcolor{green!50!black}{+10.7\%}
& 7.654 \textcolor{green!50!black}{+6.4\%}
& 8.288 \textcolor{green!50!black}{+2.2\%}
& 8.121 \textcolor{green!50!black}{+5.6\%} \\

& CITL: Best
& 8.427 \textcolor{green!50!black}{+14.7\%}
& 7.982 \textcolor{green!50!black}{+11.0\%}
& 8.704 \textcolor{green!50!black}{+7.4\%}
& 8.493 \textcolor{green!50!black}{+10.4\%} \\
\bottomrule
\end{tabular}
\end{table}

\section{Prompt Templates}
\label{app:prompts}

\subsection{Generator, $G$}

\begin{lstlisting}
You are an expert frontend web developer who is also a great UI/UX designer. You will be asked to implement a website in HTML and CSS for a given user request.

## Instructions

 - Think carefully step by step before starting the implementation
 - Create a complete HTML page for whatever the user asked you to create
 - Make sure ALL content is visible in a single screenshot - avoid modals, popups, or content that appears only on interaction
 - If interactivity is needed, ensure all interactive elements and their effects are visible simultaneously on the page
 - Use modern HTML5 semantic elements and best practices
 - Use Tailwind CSS for styling. DO NOT USE ARBITRARY VALUES (e.g. 'h-[600px]').
 - Include the Tailwind CSS CDN link in the head section, you can use: https://cdn.tailwindcss.com/
 - Make sure to use a consistent color palette.
 - Include all necessary CSS and JavaScript inline within the HTML file
 - Make sure the page is responsive and works well on different screen sizes
 - Use proper accessibility attributes and semantic HTML
 - NEVER use actual image URLs or src attributes - always create image placeholders using colored divs with appropriate Tailwind classes like 'bg-gray-300 w-32 h-32 rounded' with descriptive text inside
 - If you need icons, use simple Unicode symbols or create them with CSS/HTML
 - Make sure all functionality works without external dependencies except Tailwind CSS

## User Request

The user has requested the following:

<user_request>
{problem}
</user_request>

## Response Format

Return the full and complete HTML code for the website between ```html and ``` blocks, like this:

```html
<!DOCTYPE html>
<html lang="en">
...
</html>
```
\end{lstlisting}

\subsection{Visual Critic, $V$}

\begin{lstlisting}
You are an expert at analyzing the visual aspects of a website. You will be given a user request for a website, as well as a screenshot of the rendering of that website. Your task is to analyze the screenshot and provide feedback on how well it answers the given user query.

## Instructions

Analyze the screenshot of the website and provide feedback on the following aspects:

 - Adherance to the given user request
 - Visual layout and element positioning
 - Color scheme, contrast, and visual appeal
 - Typography, text readability, and content presentation
 - Button placement, styling, and visual feedback
 - Form design and input field appearance
 - Navigation visibility and organization
 - Content sections and visual hierarchy
 - Spacing, margins, and visual balance
 - Responsive design issues visible at current viewport
 - Placeholder image integration and visual balance
 - Missing visual elements that should be present
 - Broken or poorly styled components

Important: Focus on the design structure and layout rather than specific image content, as images will be placeholders.

## User Request

<user_request>
{problem}
</user_request>

## Response format

Start your response with an analysis of the screenshot and how well it aligns to the given user query based on the instructions given above. Then finalize your critique in this format:

<critique>
  // your comprehensive critique with suggestions for improvement
</critique>   
\end{lstlisting}

\subsection{Code Critic, $C$}

\begin{lstlisting}
You are an expert web developer tasked with analyzing the code of a website. The website is written in response to a user query, and a rendering of the website has been analyzed by an expert in visual analysis. You will be given the original user query, the code for the website, and the feedback provided by the visual analysis expert. Your task is to write a comprehensive critique of the website based on these inputs and offer suggestions for improvement.

## Instructions

- Start your response with an analysis of the given user request, website code, and expert visual feedback.
- Prioritize issues identified in the visual feedback over code quality issues, as these issues affect the experience of the end user.
- **SCREENSHOT-FIRST EVALUATION**: Focus mainly on the visual analysis feedback and how it suggests to improve the response to the user request.
- If the candidate solution imports non-existent files, note this as an issue.
- Check that all imports reference actual available libraries only if missing imports break the visual output.
- **PLACEHOLDER IMAGES**: Expect and evaluate placeholder images/divs rather than actual images - focus on their visual integration, sizing, and layout contribution.
- **IMPORTANT**: Incorporate ALL specific visual feedback issues and suggestions into your critique as top priority.
- **CRITICAL**: If the screenshot is blank or shows no content, this indicates a rendering failure - prioritize fixing what users can see.

## User Request

The original user request was as follows:

<user_request>
{problem}
</user_request>

## Solution to Evaluate

The implementation of the website is:

```html
{answer}
```

## Visual Analysis Feedback

The expert visual feedback is:

<visual_feedback>
{visual_feedback}
</visual_feedback>

## Response Format

Begin your response with an analysis of the user request, code implementation, and expert visual feedback. Then finalize your critique in this format:

<critique>
  // your comprehensive critique with suggestions for improvement
</critique>

\end{lstlisting}

\subsection{Improver, $I$}

\begin{lstlisting}
You are a web developer tasked with improving an existing website based on a provided critique. You will be given the original user request for a website, the current implementation code, and feedback with suggestions for improvements. Your task is to revise the code implementation and write a complete and improved version of the code.

## User Request

The original user request for a website is:

<user_request>
{problem}
</user_request>

## Current Implementation

The current website implementation is:

```html
{answer}
```

## Expert Critique

An expert has provided the following critique with suggestions for improvement:

<critique>
{critique}
</critique>

## Instructions:

- Carefully review the original user request, current code solution, and expert feedback given above
- Address all issues mentioned in the feedback to the best of your ability
- Use modern HTML5 semantic elements and best practices
- Use Tailwind CSS for styling via CDN. DO NOT USE ARBITRARY VALUES
- Make sure the page is responsive and works well on different screen sizes
- Use proper accessibility attributes and semantic HTML
- NEVER use actual image URLs - always create image placeholders using colored divs
- Include all necessary CSS and JavaScript inline within the HTML file
- If you need icons, use simple Unicode symbols or create them with CSS/HTML

## Response Format

Return the complete, improved HTML page in the following format:

```html
<!DOCTYPE html>
<html lang="en">
...
</html>
```
\end{lstlisting}

\subsection{Evaluator, $E$}
\label{app:evaluator}

\subsubsection{Multi-dimensional Evaluation}

\textbf{Code Judge}:

\begin{lstlisting}
You are an expert web designer and developer. Your task is to evaluate the code implementation of a webpage and determine how well it fulfills the given task requirements.

## INSTRUCTIONS

1. Analyze the task requirements to understand what needs to be accomplished
2. Examine the implementation of the solution carefully
3. Evaluate the solution against the criteria below
4. Return your response in the exact JSON format provided

## EVALUATION CRITERIA

You will be asked to provide a score of 1 to 10 on the following aspects:

- **Task accomplishment:** How well each webpage addresses the specific requirements and implicit needs. Use the following scoring guidelines:
  1-2: Major requirements missing or incorrectly implemented
  3-4: Some key requirements met but significant gaps or errors
  5-6: Most requirements addressed with minor issues or omissions
  7-8: Requirements well-implemented with only small details missing
  9-10: All requirements fully met with excellent execution

- **Code Quality:** Clarity, functionality, structure, and adherence to modern design principles
  1-2: Poor code structure, unclear logic, non-functional or buggy implementation
  3-4: Basic functionality with messy code, poor organization, or significant issues
  5-6: Working code with decent structure but room for improvement in clarity/efficiency
  7-8: Well-structured, clean code with good practices and minor areas for improvement
  9-10: Excellent code quality, clear structure, efficient implementation, best practices

## RESPONSE FORMAT

Begin your response with a thorough analysis of the code of the solution and provide an in-depth discussion on how well the solution addresses the stated task description. Focus on aspects such as task accomplishment and code quality of the solution.

Finally, summarize your final response using the following JSON format:

<summary>
{
  "task_accomplishment": {
    "summary": "<Summarize how well the solution meets the task requirements>",
    "score": "<Provide a score between 1 to 10 on how well the solution accomplishes the task>"
  },
  "code_quality": {
    "summary": "<Summarize the implementation quality of the solution>",
    "score": "<Provide a score between 1 to 10 on the code quality of the solution>"
  },
}
</summary>
\end{lstlisting}

\textbf{Visual Judge}:

\begin{lstlisting}
You are an expert web designer and developer. Your task is to evaluate the screenshot of an implemented webpage and determine how well it fulfills the given task requirements.

## INSTRUCTIONS

1. Analyze the task requirements to understand what needs to be accomplished
2. Examine the screenshot of the solution carefully
3. Evaluate the solution against the criteria below
4. Return your response in the exact JSON format provided

## EVALUATION CRITERIA

You will be asked to provide a score of 1 to 10 on the following aspects:

- **Task accomplishment:** How well each webpage addresses the specific requirements and implicit needs. Use the following scoring guidelines:
  1-2: Major requirements missing or incorrectly implemented
  3-4: Some key requirements met but significant gaps or errors
  5-6: Most requirements addressed with minor issues or omissions
  7-8: Requirements well-implemented with only small details missing
  9-10: All requirements fully met with excellent execution

- **Aesthetic quality:** Visual appeal, layout, typography, color usage, and modern design principles
  1-2: Poor visual design, cluttered layout, poor color/typography choices
  3-4: Basic design with noticeable visual issues or outdated appearance
  5-6: Decent design with some visual appeal but room for improvement
  7-8: Good visual design with modern aesthetics and clean layout
  9-10: Excellent visual design, professional appearance, outstanding aesthetics

## RESPONSE FORMAT

Begin your response with a thorough analysis of the screenshot of the solution and provide an in-depth discussion on how well the solution addresses the stated task description. Focus on aspects such as task accomplishment and aesthetic qualities of the solution.

Finally, summarize your final response using the following JSON format:

<summary>
{
  "task_accomplishment": {
    "summary": "<Summarize how well the solution meets the task requirements>",
    "score": "<Provide a score between 1 to 10 on how well the solution accomplishes the task>"
  },
  "aesthetic_quality": {
    "summary": "<Summarize the aesthetic qualities of the solution>",
    "score": "<Provide a score between 1 to 10 on the aesthetic quality of the solution>"
  },
}
</summary>

\end{lstlisting}

\subsubsection{Single-dimensional Evaluation}
\label{subsub:single_dim}
\begin{lstlisting}
You are a meticulous code reviewer and design analyst. A user has provided a request for a website. You will be given the HTML for the website and a screenshot of the rendered HTML code, and your role is to evaluate the quality based on code correctness, structural integrity, visual rendering, stylistic coherence, and adherence to the user request. Be sure to be CRITICAL!

## Task Instructions

1. Analyze the results in depth:
   * Begin with a long, structured analysis of the HTML code and the associated screenshot.
     - Focus especially on the screenshot of the rendered code:
       + Is the HTML code rendered correctly?
       + Are all visual elements visible, aligned, and styled appropriately?
       + Does it clearly align with the problem description?
       + Is the layout balanced and visually appealing?
     - Consider code quality aspects:
       + Are there any major syntax or structural problems?
       + Are styles consistent and semantically appropriate?
       + Is the JavaScript code clean and well-structured?

2. Assign a final score using the 10-point scale based on the total evaluation using the following scoring guidelines.

## Scoring Guidelines

* 1 = Completely Broken
  Non-functional. Major rendering failures. Completely broken layout. Unreadable. No adherence to problem description.

* 2 = Very Poor
  Mostly non-functional. Major rendering failures. Broken layout. Poor adherence to problem description.

* 3 = Poor
  Mostly broken visuals. Major code/styling issues. Incoherent or hard to follow. Poor adherence to problem description.

* 4 = Below Average
  Somewhat functional but very unpolished. Multiple significant issues in code or layout. Many parts of the problem description not implemented.

* 5 = Fair
  Basic functionality but unpolished. Several issues in code or layout. Some parts of the problem description not implemented.

* 6 = Acceptable
  Mostly functional. Minor bugs or inconsistencies. Mediocre visual appeal. Most basic requirements met but lacks polish.

* 7 = Good
  Functional and coherent. Minor code or style issues. Clear layout. Most parts of problem description implemented well.

* 8 = Very Good
  High-quality result. Clean visual presentation. Polished and intentional. Most parts of the problem description implemented excellently.

* 9 = Excellent
  Near-perfect implementation. Minimal issues. Visually compelling and technically sound. Covers all points in the problem description with high quality.

* 10 = Perfect
  Flawless implementation. No issues. Exceptional visual design and code quality. Exceeds all requirements in the problem description.

## User Request

<user_request>
{problem}
</user_request>

## HTML Code to Evaluate

```html
{answer}
```

## Final Output

After completing your analysis:

* Output an in-depth review based on the evaluated code and the visual rendering between <review> and </review> tags
* Then, output your score as a number between 1 and 10 between <score> and </score> tags.

Do not include the score until after the full analysis has been completed.

\end{lstlisting}

\subsection{Refinement without critique}
\label{app:refine_wo_crtic}

\begin{lstlisting}
You are an HTML development expert improving an existing web page. 

## Input:
- Original Requirements: {problem}
- Current Solution: {answer}

## Instructions:
1. Improve layout, styling, and user experience.
2. Maintain functionality while enhancing code quality.
3. Ensure responsive design with Tailwind CSS.
4. CRITICAL: Use inline CSS/JavaScript only - no external files
5. NEVER use actual image URLs or src attributes - always create image placeholders using colored divs
6. All code must be completely self-contained HTML

## Response Format

Return the full and complete HTML code for the website between ```html and ``` blocks, like this:

```html
<!DOCTYPE html>
<html lang="en">
...
</html>
```
\end{lstlisting}

\subsection{critique classification}
We first used a sample of 50 LLM critiques and asked Claude 4.5 Sonnet to identify the common themes or categories of these critiques. Next, we use these 8 categories to classify a larger dataset of  critiques.
\label{app:critique_classification}
\begin{lstlisting}
You are an expert on web design, your task is to match the following comments and critiques about web design into 8 predefined categories based on what these comments and critiques are about.

The 8 categories are:

1. Visual Design & Polish : Visual aesthetics, such as poor padding and margins, contrasts
2. Interactivity & Feedback : No loading/status indicators, user actions don't provide clear confirmation, no indication of current page/section in navigation
3. Missing Critical Elements : Incomplete sections, No progress indicators, absent metadata, no search/filter functionality
4. Typography & Readability : Text too small, Inconsistent font hierarchy, Poor line height
5. Component Design : Weak card styling, inconsistent buttons, generic/small icons
6. Responsive Design : Designs don't adapt well to smaller screens, elements don't stack or reflow properly
7. Content & Structure : Large areas with minimal content, content and elements not logically organized or appear connected
8. Implementation Gaps : Requested features not implemented, accessibility concerns like lacking of keyboard navigation

Please output only using the category name above, if you think the critique belongs to more than 1 category, pick only the most relevant one as the category.

The critique text is:

{critique_text}

Always start your answer with the word "Category:"

Your answer:

\end{lstlisting}

\end{document}